%% file: main.tex
\documentclass[letterpaper]{article} 
\usepackage{aaai24}  
\usepackage{times}  
\usepackage{helvet} 
\usepackage{courier} 
\usepackage[hyphens]{url} 
\usepackage{graphicx} 
\usepackage{natbib}
\usepackage{caption}
\usepackage{algorithm}
\usepackage{algorithmic}

\usepackage{amsmath}
\usepackage{amssymb}
\usepackage{booktabs}
\usepackage{makecell}
\usepackage{multirow}
\input{math_commands.tex}

\usepackage{marvosym}
\newcommand{\etal}{\emph{et al}.}

\usepackage{newfloat}
\usepackage{listings}
\DeclareCaptionStyle{ruled}{labelfont=normalfont,labelsep=colon,strut=off} 
\floatstyle{ruled}
\newfloat{listing}{tb}{lst}{}
\floatname{listing}{Listing}

\title{Hand-Centric Motion Refinement for 3D Hand-Object Interaction via Hierarchical Spatial-Temporal Modeling}
\author {
    Yuze Hao\textsuperscript{\rm 1,\rm 2, \thanks{This work was done when Yuze Hao was an intern at Zhejiang University. The code is publicly available at https://github.com/Holiday888/HST-Net.}},
    Jianrong Zhang\textsuperscript{\rm 1},
    Tao Zhuo\textsuperscript{\rm 3},
    Fuan Wen\textsuperscript{\rm 2,4},
    Hehe Fan\textsuperscript{\rm 1, \thanks{Corresponding author.}}
}
\affiliations {
    \textsuperscript{\rm 1}Zhejiang University\\
    \textsuperscript{\rm 2}Beijing University of Posts and Telecommunications\\
    \textsuperscript{\rm 3}Shandong Artificial Intelligence Institute, Qilu University of Technology (Shandong Academy of Sciences)\\
    \textsuperscript{\rm 4}Beijing Key Laboratory of Network System and Network Culture

}

\usepackage{bibentry}

\begin{document}

\maketitle

\begin{abstract}
Hands are the main medium when people interact with the world. Generating proper 3D motion for hand-object interaction is vital for applications such as virtual reality and robotics. Although grasp tracking or object manipulation synthesis can produce coarse hand motion, this kind of motion is inevitably noisy and full of jitter. To address this problem, we propose a data-driven method for coarse motion refinement. First, we design a hand-centric representation to describe the dynamic spatial-temporal relation between hands and objects. Compared to the object-centric representation, our hand-centric representation is straightforward and does not require an ambiguous projection process that converts object-based prediction into hand motion. Second, to capture the dynamic clues of hand-object interaction, we propose a new architecture that models the spatial and temporal structure in a hierarchical manner. Extensive experiments demonstrate that our method outperforms previous methods by a noticeable margin. 

\end{abstract}

\section{Introduction}
Recently, hand-object interaction modeling has been widely used in many applications, such as VR/AR~\cite{holl2018efficient,canales2019virtual}, and robotics~\cite{xu2023unidexgrasp,qin2023dexpoint,bao2023dexart}. Recent methods can directly predict the hand-object pose from images~\cite{chen2021joint,doosti2020hope} or generate hand grasping pose given object information~\cite{jiang2021hand,brahmbhatt2019contactgrasp,zheng2023cams}. However, due to the high degrees of freedom of hand articulation, self-occlusion of hands, and mutual occlusion of objects, the predicted results tend to be error-prone, limiting its application in downstream tasks.

%--------------------------------------------Teaser1----------------------------------------------
\begin{figure}[!ht]
\centering
\includegraphics[width=0.45\textwidth]{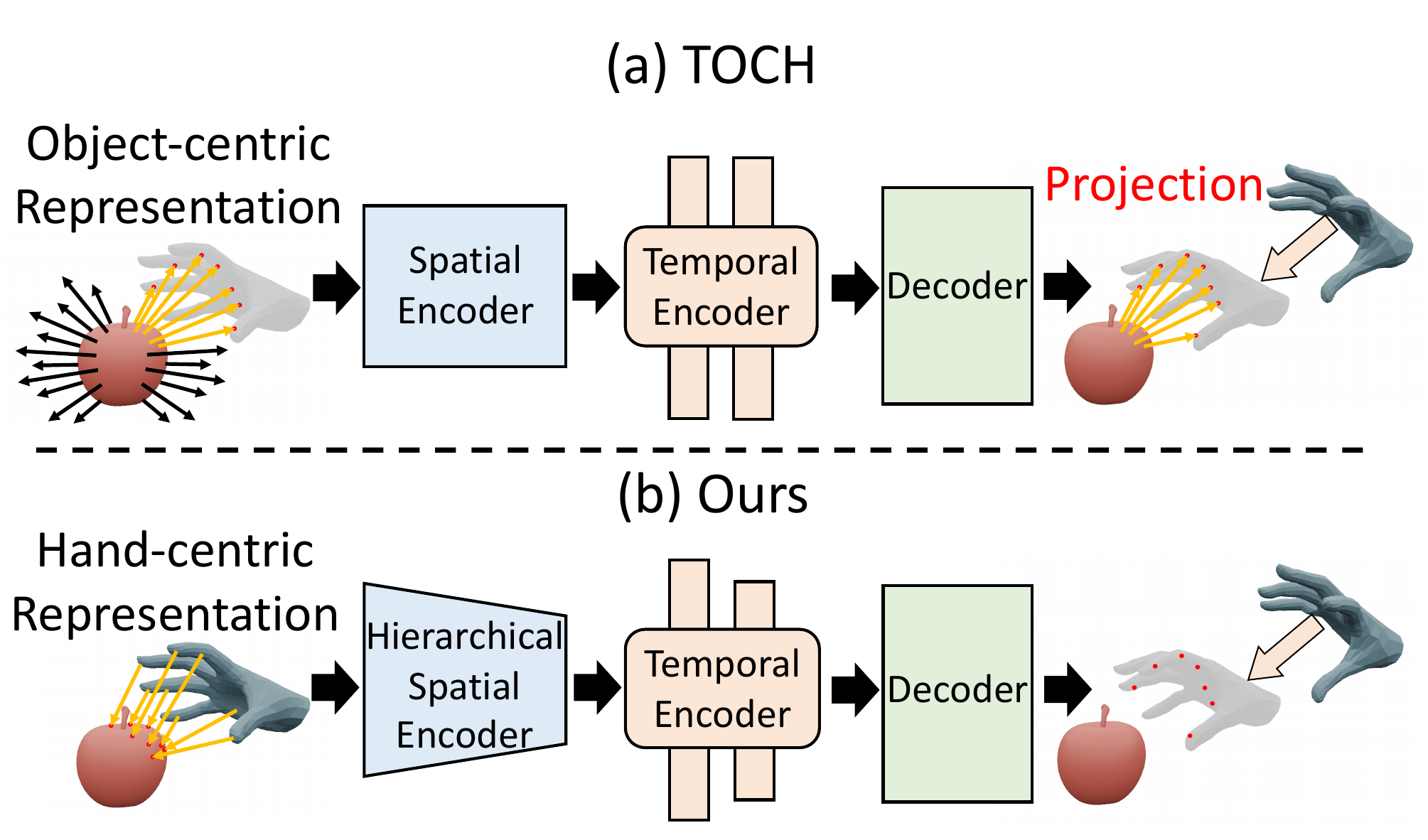}
\caption{Comparing with TOCH~\cite{zhou2022toch}, our method has three advantages. (1) To capture the relation between hand and object, the existing method first emits rays from the object and then collects points that arrive at the hand. In contrast, we propose a straightforward hand-centric representation, which directly models the hand-object interaction. (2) Our hierarchical spatial-temporal architecture better captures dynamic information across different scales than the fixed-scale design in TOCH. (3) Due to the direct hand-object representation, our method does not require the additional post-process that converts object-centric representation into hand motions.}
\label{fig:intro}
\end{figure}
%------------------------------------------------------------------------------------------

Many approaches~\cite{grady2021contactopt,tse2022s} are proposed to solve the aforementioned problems. To alleviate the potential anatomical irregularities in hand poses, Yang~\etal~\cite{yang2021cpf} introduced a joint bending constraint to the parametric hand model~\cite{romero2017embodied}. To avoid the intersection between hands and objects or noncontact grasping, a heuristic repulsion loss and an attraction loss are employed~\cite{hasson19_obman}. 
However, these methods mainly consider the contacting moment (a single frame), instead of the entire hand-object interaction process (multiple frames) that includes both approaching and contacting stages. 

In contrast, hand-object tracking, reconstruction~\cite{hasson20_handobjectconsist,hasson20_handobject} and interaction synthesis~\cite{taheri2022goal,wu2022saga} can produce hand-object interaction motions, which involves pre-grasping and post-manipulation stages. Although these methods extend hand-object interaction from the contacting moment to the entire process, they introduce the jitter issue and inconsistent contact, leading to perturbed hand-object interaction motions. 
To refine the dynamic hand-object interaction motions, TOCH~\cite{zhou2022toch} employs PointNet~\cite{qi2017pointnet} for spatial modeling and Gated Recurrent Unit (GRU)~\cite{chung2014empirical} for temporal modeling.  
However, this fixed-scale spatial-temporal architecture is not efficient in capturing the spatial-temporal structure from local to global, thus limiting the ability of deep neural networks to learn the fine-coarse representations for hand-object interaction. 
Besides, as hand-object interaction is highly correlated with the object, TOCH employs an object-centric representation strategy that denotes hand motions from the object perspective, as shown in Figure~\ref{fig:intro}(a). However, object-centric representation suffers from low-level information utilization because only a partial surface region of the object interacts with the hand. Moreover, employing such a representation necessitates an additional projection process, potentially introducing more ambiguity into the outcomes. To be specific, given the refined hand motions represented in the object-centric fashion, a projection process is required to align the hand pose with its corresponding representation. Nonetheless, when the contact area between the hand and object is small, it is difficult for the object-centric representation to correctly reflect the hand pose. 

In this paper, we first introduce a novel hand-centric representation, based on hand poses and motions, as illustrated in Figure~\ref{fig:intro}(b).  
The proposed representation captures the hand-object correspondence with the displacement between each hand vertex and the anchor points of the object. 
Compared to the previous object-centric method~\cite{zhou2022toch}, the hand-centric representation does not require information from the untouched, nontarget, or meaningless object areas.  
Moreover, our method is straightforward and eliminates the need for an additional projection process. 
Second, the previous fixed-scale architecture overlooks granularity in region-level contact, global-level pose, and short-term to long-term temporal dynamics. 
To address this issue, we introduce a hierarchical spatial-temporal encoder. 
The spatial encoder integrates regional details for a broader receptive field.
The temporal encoder employs a transformer architecture across multiple frames, capturing both short and long-term dependencies.  Extensive experiments on the GRAB and HO3D datasets demonstrate the effectiveness of the proposed method. 
In summary, our contributions are three folds:
\begin{itemize}
    \item 
   
    We propose a new hand-centric representation that effectively utilizes the hand-object interaction information. This representation avoids the ambiguity problem from the post-projection process.

    \item We propose a hierarchical spatial-temporal architecture to model hand-object interaction from the local level to the global level, which effectively captures the detailed clues and overall information of interactions.
    \item Experimental results demonstrate that our method outperforms the existing methods across various metrics, especially for longer-distant interaction sequences.
\end{itemize}

%------------------------------------------- Architecture Figure-----------------------------------------------------------------
\begin{figure*}[ht]
\centering
\includegraphics[width=0.95\textwidth]{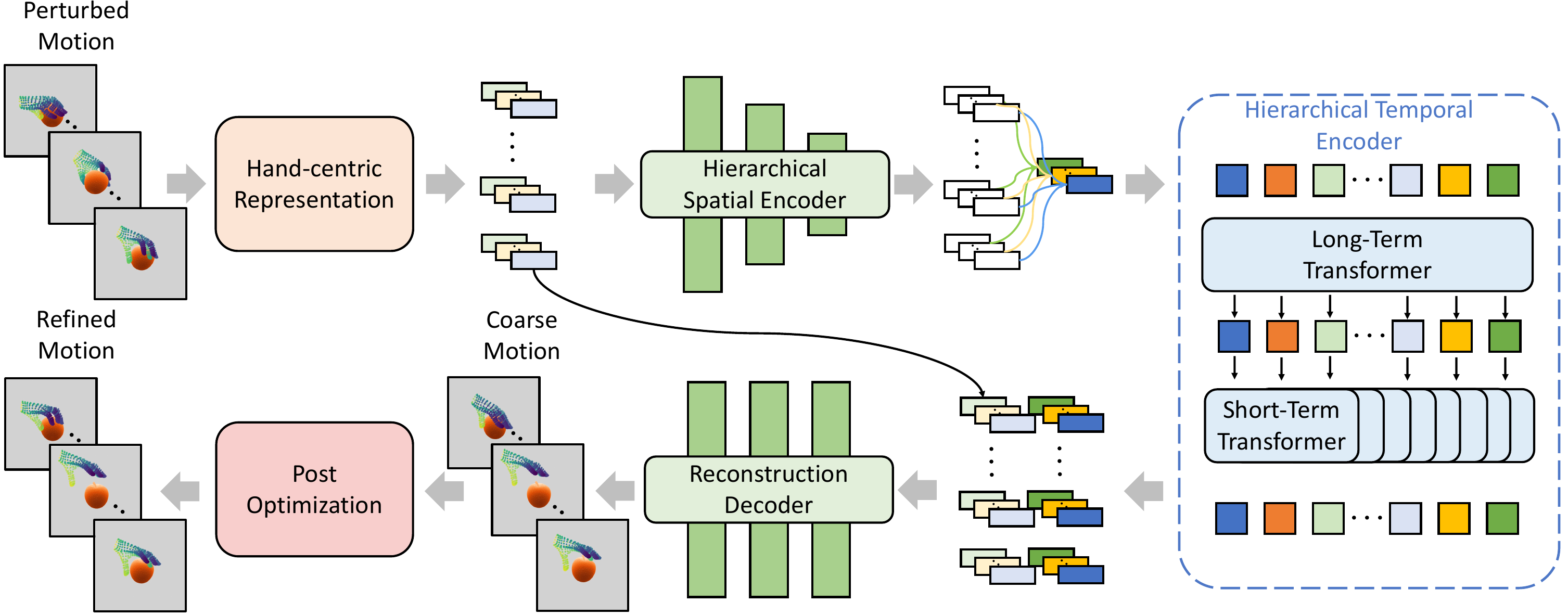}
\caption{Overview of our framework.  Given a perturbed interaction sequence, we first convert the sequence into our hand-centric correspondence representation. Then the representations are fed into a hierarchical spatial encoder to capture the local-global spatial features for each frame. Next, the features are passed through a hierarchical temporal encoder to extract long-term and short-term dependencies across frames. Lastly, the refined sequences are obtained from a reconstruction decoder followed by a post-optimization.}
\label{fig:architecture}
\end{figure*}
%----------------------------------------------------------------------------------------------------------------------------------

%----------------------------------------Related Work-------------------------------------
\section{Related Work}
\subsection{Hand Object Interaction}
Hand-object interaction is important in virtual reality and robotics. Recently, with the advent of datasets that contain both hand and object annotations~\cite{hampali2020honnotate,taheri2020grab,hasson19_obman,fan2023arctic}, impressive progress has been made.

Early methods mainly focus on modeling static grasping interaction. Some approaches~\cite{lin2023harmonious,doosti2020hope,hasson19_obman} directly estimate hand-object pose or reconstruct 3D mesh from a single image. Given the information of objects, several efforts~\cite{jiang2021hand,brahmbhatt2019contactgrasp,turpin2022grasp} generate the corresponding hand-grasping pose. Furthermore, other approaches focus on refining hand-object grasping state~\cite{grady2021contactopt,yang2021cpf,tse2022s,taheri2020grab}. For instance, Grady~\etal~\cite{grady2021contactopt} proposed to predict the hand-object contact map and then align the hand with the predicted region. However, these methods only consider the static grasping stage. 

Recently, there has been a growing emphasis on modeling dynamic hand-object interactions. Some methods~\cite{hasson20_handobjectconsist,hasson20_handobject,chen2023tracking} can estimate/reconstruct interactions from video input. With the hand-object interaction motion datasets~\cite{taheri2020grab,hampali2020honnotate,liu2022hoi4d,fan2023arctic}, interaction motions can be generated~\cite{zhang2021manipnet,taheri2022goal,wu2022saga,zheng2023cams}. As the first method to address refining perturbed 3D hand-object interaction sequence, Zhou~\etal~\cite{zhou2022toch} proposed an object-centric representation with a spatial-temporal modeling architecture. In this work, we share the same goal to refine the hand-object interaction sequence.

\subsection{Modeling Human Motion Prior}
Obtaining high-quality global motion prior is crucial for human motion modeling, by capturing the motion prior, tasks such as motion prediction~\cite{zhao2023bidirectional,xu2022diverse} and motion generation~\cite{zhang2023t2mgpt,tevet2022human,guo2022tm2t} can be benefited. Researchers have explored various approaches to obtain spatial-temporal motion prior through auto-encoder~\cite{fragkiadaki2015recurrent, rempe2021humor}, generative adversarial network~\cite{ferreira2021learning,zhao2020bayesian}, and diffusion model~\cite{zhang2022motiondiffuse,tevet2022human}.

For hand motion, similar to body motion modeling, by capturing the spatial-temporal motion prior, tasks like hand gesture prediction~\cite{qi2023diverse,ng2021body2hands}, dynamic hand pose estimation~\cite{liu2022spatial} can be achieved. In terms of hand-object interaction, modeling interaction sequences in a spatial-temporal manner~\cite{ zhou2022toch} is also essential. In this work, we handle spatial-temporal information of hand-object interaction hierarchically.

\subsection{Point Cloud Video Modeling}
Hand-object interactions are represented as sequences of 3D point clouds or point cloud videos. While various studies have focused on modeling dynamic point clouds~\cite{fan2021point,liu2022hoi4d,fan2022pstnet,shen2023pointcmp,fan2022point,sheng2023contrastive,fan2021deep,liu2023leaf} for classification and segmentation, our approach can be regarded as a point cloud video denoising method.

%----------------------------------------Methodology-------------------------------------
\section{Methodology}
\subsection{Overview and Problem Setup}
Our framework is illustrated in Figure~\ref{fig:architecture}. Given a perturbed hand sequence with corresponding object sequence $([\tilde{\mH}^0;\mO^0], [\tilde{\mH}^1;\mO^1], \cdots, [\tilde{\mH}^T;\mO^T])$ where $T$ is the number of frames for interaction sequence. For each frame $t$, $\tilde{\mH}^t=\{\tilde{\vh}_i^t\}_{i=1}^{N^h}$ and $\mO^t=\{\vo_i^t\}_{i=1}^{N^o}$ are point cloud representations for the hand and object, respectively. Besides, $\tilde{\vh}_i^t \in \mathbb{R}^3 $ and $\vo_i^t \in \mathbb{R}^3$ are the $i$-th vertex for hand and object, and $N^h$ and $N^o$ are the respective point counts. Note that $\tilde{\mH}^t$ is the hand mesh vertices that are generated from MANO \cite{romero2017embodied}. Our goal is to refine the perturbed hand sequence and generate a more realistic hand interaction sequence $\hat{\tH} = [\hat{\mH}^0, \hat{\mH}^1, \cdots, \hat{\mH}^T]$ where $\hat{\mH}^t \in \mathbb{R}^{{N^h} \times 3}$. As we focus on enhancing the perturbed hand motions, we follow previous work~\cite{zhou2022toch} to assume the ground truth object sequences $\tO = [\mO^0, \mO^1, \cdots, \mO^T]$ are available during the training and testing stage.

To refine the perturbed hand motions, we first convert the perturbed hand-object interaction sequence pair into our hand-centric representation. Then we feed the motion representation into the designed hierarchical spatial-temporal architecture, generating coarse motions. Finally, a post-optimization process is applied to further refine the artifacts and smoothness.

\subsection{Hand-Centric Representation}
\label{hand-centric representatioin}
TOCH~\cite{zhou2022toch} represents the hand-object interaction in an object-centric manner. This is achieved by emitting rays from the vertices of the object and subsequently gathering points that intersect with the hand. TOCH utilizes both the distance of the ray and the positional information on the hand as distinctive features for interaction. However, such a representation method suffers from inefficient data utilization, because only a portion region of the object interacts with the hand (as shown in Figure~\ref{fig:intro}(a)). Furthermore, the representation needs an additional projection process, potentially introducing more ambiguity into the outcomes, such as the visual results shown in Figure~\ref{fig:visualization1}.

In this work, we propose a hand-centric representation to tackle the above problems. Specifically, for each frame in the sequence, given the perturbed hand $\tilde{\mH}^t$ and object $\mO^t$. The corresponding representation is defined as follows: 
\begin{equation}
   \mR^t =  \{(\vec{\vc_i^t},\tilde{\vh}_i^t)\}_{i=1}^{N^h},
\label{formula:handcentricfeature}
\end{equation}
where $\vec{\vc_i^t} = \hat{\vo}_i^t - \tilde{\vh}_i^t$, $\hat{\vo}_i^t \in \mO^t$ is the closest vertex to $\tilde{\vh}_i^t$. Based on such a hand-centric representation, our model integrates the relevant object information while effectively preserving the distinct characteristics of each hand vertex. Subsequently, the representation of the hand-object sequence is defined as $\tR = [\mR^0, \mR^1, \cdots, \mR^T]$ with $\mR^t = \{\vr_i^t\}_{i=1}^{N^h}  \in \mathbb{R}^{{N^h} \times 6}$.

\subsection{Hierarchical Spatial-Temporal Motion Reconstruction}
\label{spatial-temporal hierarchical autoencoder}
Unlike the previous method~\cite{zhou2022toch} that focuses on fixed-scale hand-object interaction point cloud sequences, our approach employs a Hierarchical Spatial-Temporal Network (HST-Net). This network comprises hierarchical spatial and temporal encoders, along with a reconstruction decoder. Through HST-Net, our model captures region-level contact, global-level pose, and long-term and short-term dependencies, leading to a more comprehensive abstraction of the interaction sequence.

\paragraph{Hierarchical spatial encoder.} 
Given the sequence $\tR = [\mR^0, \mR^1, \cdots, \mR^T]$, the proposed hierarchical spatial encoder encodes the sequence to $\mS=[\vs^0, \vs^1, \cdots, \vs^T]$, where $\mS \in \mathbb{R}^{d \times T}$ is the latent representation of the sequence with $d$ is the dimension of the latent feature.

As our representation is a set of points attached to its relative feature, we propose to model the feature in a hierarchical local-global manner. Specifically,  given the point cloud $\tilde{\mH}^t$ with the corresponding feature $\mR^t$. To get local level information, we first select several anchor points $\tilde{\mH}^{\prime t}$ from $\tilde{\mH}^t$ by using farthest point sampling. With the selected anchors, a continuous convolution layer is used to capture the local spatial information for each point:

\begin{equation}
    \vf^t_{(x,y,z)} = \underset{\left \| \delta_x, \delta_y, \delta_z  \right \| \le \gamma} {\texttt{MAX}} \ \texttt{MLP}(\vr^t_{(x+\delta_x, y+\delta_y, z+\delta_z)}, \delta_x, \delta_y, \delta_z),
\label{formula:spatial-encoding}
\end{equation}
where $\vf^t_{(x,y,z)}$ is the corresponding spatial feature of point $(x,y,z)$ from $\tilde{\mH}^{\prime t}$, and $(\delta_x, \delta_y, \delta_z)$ represents the displacement, $\vr^t_ {(x+\delta_x, y+\delta_y, z+\delta_z)}$ is the feature of position $(x+\delta_x, y+\delta_y, z+\delta_z)$ at time $t$, and $\gamma$ is the spatial search radius. 

After multiple layers of sampling and aggregation, the local receptive field is sufficiently enlarged. Then the global feature is obtained by applying the max-pooling operation on the output of the final spatial layer:
\begin{equation}
    \vs^t = \underset{(x,y,z) \in {\tilde{\mH}^{\prime t}}} {\texttt{MAX}} \ {\vf^t_{(x,y,z)}}.
\label{formula:maxpooling}
\end{equation}

\paragraph{Hierarchical temporal encoder.}
Based on the frame-wise spatial feature, we propose to use long-term and short-term transformers to obtain the inter-frame hierarchical dependencies. The former performs long-range information integration across all temporal dimensions, and the latter focuses more on the relationship within the local receptive field. Precisely, given the spatial features $\mS=[\vs^0, \vs^1, \cdots, \vs^T]$, we first take transformers with long-term self-attention to get global information which can be formulated as: 
\begin{equation}
    \mL = \texttt{Self-Attention}\left(\vs^0, \vs^1, \cdots, \vs^T\right),
\end{equation}
where $\mL = [\vl^0, \vl^1, \cdots, \vl^T ]$ and $\mL \in \mathbb{R}^{d' \times T}$. Then the long-term representation $\mL$ is split into $B$ frame bins, obtaining $\hat{\mL} = \{\hat{\mL}_b\}_{b=1}^B$, where $\hat{\mL}_b =[\hat{\vl}_b^0, \hat{\vl}_b^1, \cdots, \hat{\vl}_b^{T'}]$ denotes a frame bin with $T' = T/B$ frames. Then short-term self-attention is employed for local feature integration, the process can be computed as follows:  
\begin{equation}
    \tilde{\mS}_b = \texttt{Self-Attention}(\hat{\vl}_b^0, \hat{\vl}_b^1, \cdots, \hat{\vl}_b^{T'}),
\end{equation}
where $\tilde{\mS}_b = [\tilde{\vs}_b^0, \tilde{\vs}_b^1, \cdots, \tilde{\vs}_b^{T'}]$ and $\tilde{\mS}_b \in \mathbb{R}^{d'' \times T'}$. Finally, we concatenate $B$ short-term representations for $\tilde{\mS} = [\tilde{\vs}^0, \tilde{\vs}^1, \cdots, \tilde{\vs}^T]$ as the input of the decoder, where $\tilde{\mS} \in \mathbb{R}^{d'' \times T}$.

\paragraph{Reconstruction decoder.}
Given the frame-wise global feature $\tilde{\mS}$ obtained by our hierarchical spatial-temporal encoder, the reconstruction decoder is employed to recover coarse hand motions $\hat{\tH}'$. The feature is first repeated for each point and then concatenated with the hand-centric representation, which can be formulated as $\tilde{\mR}^t = \texttt{Concat}(\tilde{\mS}^t, \mR^t)$, where $\tilde{\mR}^t \in \mathbb{R}^{{N^h} \times {(d''+6)}}$. Then, based on the input $[\tilde{\mR}^0, \tilde{\mR}^1, \cdots, \tilde{\mR}^T]$, we use several weight-shared MLPs as the decoder to reconstruct hand motion $\hat{\tH}'$.

Given the ground truth of hand motion $\tH = [\mH^0, \mH^1, \cdots, \mH^T]$ with $\mH^t \in \mathbb{R}^{{N^h} \times 3}$. We minimize the distance between ground truth $\tH$ and prediction $\hat{\tH}'$ during training, which can be written as:

\begin{equation}
   \mathcal{L}_{recons} = \left \| \tH - \hat{\tH}' \right \|_2^2 .
\end{equation}

\subsection{Post Optimization}
Based on the predicted coarse hand motion $\hat{\tH}'$, we further employ a post-optimization step to improve the overall smoothness. We use the MANO model~\cite{romero2017embodied} with predicted hand vertices, to generate final refined hand motion $\hat{\tH}$. Here, the MANO model serves as a linear blend skinning model that can produce the 3D hand vertices based on the shape and pose parameters. We optimize the shape parameter $\beta$ , pose, orientation and translation parameter $\theta= \{\theta^t\}_{t=1}^T$ of the MANO model by minimizing:
\begin{equation}
    \mathcal{L}(\beta,\theta)= \mathcal{L}_{consistent}(\beta,\theta)+\mathcal{L}_{smooth}(\beta,\theta),
\end{equation}
where $\mathcal{L}_{consistent}$ ensures alignment with the predicted outcomes $\hat{\tH}'$ to maintain consistency, and $\mathcal{L}_{smooth}$ aims to enhance overall smoothness. Importantly, we assume that the shape remains constant throughout the given interaction sequence, allowing us to utilize the same shape parameter $\beta$ across the sequence.

%----------------------------------------Experinment-------------------------------------
\begin{table*}[t]   
\small
\centering\setlength{\tabcolsep}{16.2pt}
\begin{tabular}{c|c|c|c|c|c}
\toprule
\multicolumn{2}{c|}{Methods} & MPJPE (mm) $\downarrow$ &  MPVPE (mm) $\downarrow$ & IV ($cm^3$) $\downarrow$ &  C-IoU (\%) $\uparrow$  \\
\midrule

\multirow{3}*{\makecell{Task 1:GRAB-T \\ (0.01)}} & Perturbation & 16.0   & 16.0 & 2.48 & 16.24 \\
\cmidrule{2-6}
~ & TOCH & 9.93 & 11.8 & 1.79 & \hspace{0.5em}23.25* \\
~ & Ours & \textbf{6.90} & \textbf{7.75} & \textbf{1.70} & \textbf{28.00} \\
\midrule

\multirow{3}*{\makecell{Task 2: GRAB-T \\ (0.02)}} &Perturbation & 31.9 & 31.9 & 2.4 & 10.69 \\
\cmidrule{2-6}
~ & TOCH & 12.3 & 13.9 & 2.5 & \hspace{0.5em}20.35* \\
~ & Ours & \textbf{10.19} & \textbf{10.76} & \textbf{1.85} & \textbf{22.84} \\
\midrule

\multirow{3}*{\makecell{Task 3: GRAB-R \\ (0.3)}} & Perturbation & 4.58 & 6.30 & 1.88 & 23.21\\
\cmidrule{2-6}
~ & TOCH & 9.58 & 11.5 & \textbf{1.52} & \hspace{0.5em}23.39*\\
~ & Ours & \textbf{5.85} & \textbf{6.92} & 1.61 & \textbf{28.61} \\
\midrule

\multirow{3}*{\makecell{Task 4: GRAB-R \\ (0.5)}} & Perturbation & 7.53 & 10.3 & 1.78 & 17.31\\
\cmidrule{2-6}
~ & TOCH & 9.12 & 11.0 & \textbf{1.35} & \hspace{0.5em}22.13* \\
~ & Ours & \textbf{6.28} & \textbf{7.54} & 1.69 & \textbf{25.27} \\
\midrule

\multirow{3}*{\makecell{Task 5: GRAB-B \\ (0.01 \& 0.3)}} & Perturbation & 17.3 & 18.3 & 2.20 & 12.14 \\
\cmidrule{2-6}
~ & TOCH  & 10.3 & 12.1 & 1.78 & \hspace{0.5em}23.10* \\
~ & Ours & \textbf{7.17} & \textbf{8.09} & \textbf{1.69} & \textbf{26.36} \\
\bottomrule
\end{tabular}
\caption{Quantitative results on the GRAB~\cite{taheri2020grab} test set. Following TOCH~\cite{zhou2022toch}, we manually perturb the ground truth with GRAB-T (translation-dominant perturbation), GRAB-R (pose-dominant perturbation), and GRAB-B (balanced perturbation). The numbers inside the parentheses represent the magnitude of perturbation which is sampled from Gaussian noise. The symbol $*$ denotes that we reproduced the results using the released code of TOCH~\cite{zhou2022toch}.}
\label{tab:GRAB}
\end{table*}

\begin{table}[h]
\small
\centering\setlength{\tabcolsep}{8.5pt}
\begin{tabular}{c|l|c|c|c}
\toprule
\multicolumn{2}{c|}{Methods} & MPJPE &  MPVPE & IV\\ \midrule
Tracking & Hasson~\etal & 11.4 & 11.4 & 9.26 \\ \midrule
\multirow{2}*{Static} & RefineNet & 11.6 & 11.5 & 8.11 \\ 
~ & ContactOpt & 9.47 & 9.45 & 5.71 \\  \midrule
\multirow{2}*{Dynamic} & TOCH & 9.32 & 9.28 & 4.66 \\
~ & Ours & \textbf{9.18} & \textbf{9.21} & \textbf{4.52} \\
\bottomrule
\end{tabular}
\caption{Quantitative evaluation on the HO3D dataset. In this experiment, we follow the method in TOCH~\cite{zhou2022toch} to select interaction sequences from the predicted result of the tracking method~\cite{hasson20_handobjectconsist} and use different optimization methods~\cite{taheri2020grab,grady2021contactopt} to refine interaction sequences.}
\label{tab:tracking}
\end{table}

%------------------------------------------- Visiualization-----------------------------------------------------------------
\begin{figure*}[t]
\centering
\includegraphics[width=0.9\textwidth]{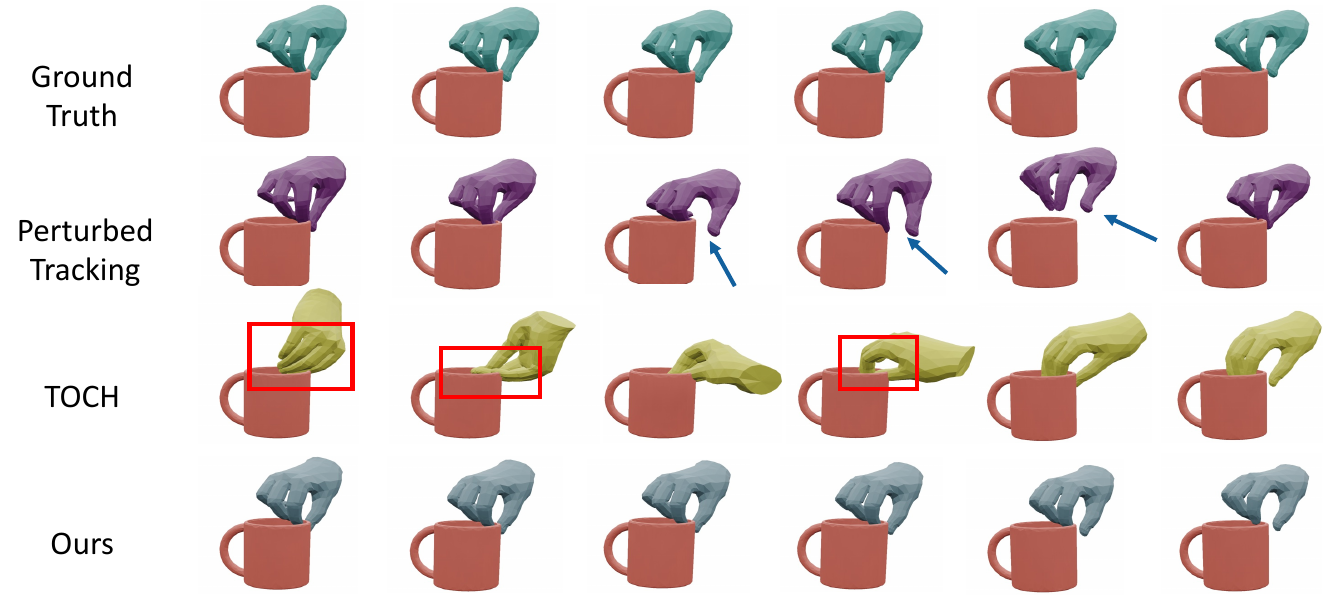}
\caption{Qualitative results for refining inconsistency pose (highlighted with blue arrows) in perturbed tracking sequence. The refined sequence of TOCH~\cite{zhou2022toch} exhibits improper grasping poses (highlighted with red boxes). In contrast, our reconstructions demonstrate a more plausible interacting pose.}
\label{fig:visualization1}
\end{figure*}
%------------------------------------------------------------------------------------------------------------

%------------------------------------------- Visiualization-----------------------------------------------------------------
\begin{figure*}[t]
\centering
\includegraphics[width=0.9\textwidth]{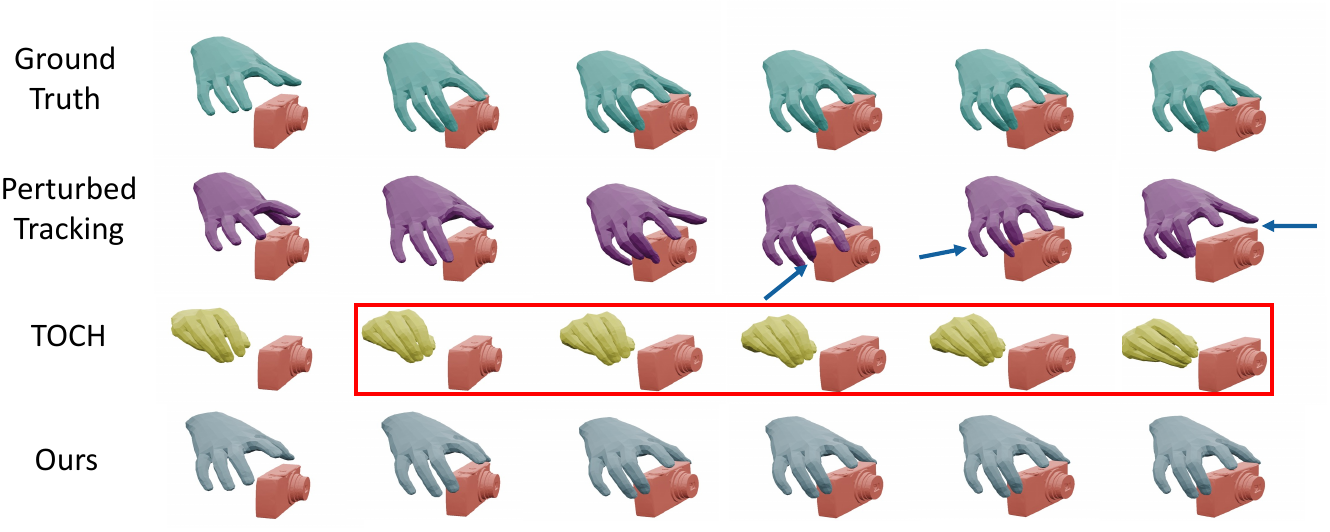}
\caption{Qualitative results for refining inter-penetration (highlighted with blue arrows) in perturbed tracking sequence. The refined sequence of TOCH~\cite{zhou2022toch} exhibits inadequate contact (highlighted with red boxes) while our results can achieve more realistic interaction.}
\label{fig:visualization2}
\end{figure*}
%------------------------------------------------------------------------------------------------------------
\section{Experiments}
In this section, we apply our method to synthetic and real tracking errors for hand-object interaction and evaluate our method on various metrics. We first introduce the experiment settings including the datasets, evaluation metrics, and implementation details. Then we demonstrate that our method could achieve better performance compared with the state-of-the-art methods. Finally, we demonstrate the effectiveness of the proposed components with ablation studies.

\subsection{Dataset}
\label{Dataset}
\paragraph{GRAB.}

GRAB~\cite{taheri2020grab} is a Motion Capture dataset focusing on whole-body grasping. 
It comprises complete 3D shape and pose sequences, including 10 performers interacting with 51 objects.
Following TOCH~\cite{zhou2022toch}, we select 47/4/6 objects for training, validation, and testing, and filter out frames where the hand wrist is more than 15cm away from the object. 
To simplify the problem, we use the right hand as the case of study.
For left-hand cases, we mirror hands and objects. 
To evaluate our method on synthetic perturbations, we follow TOCH and adopt various perturbation strategies. These strategies include GRAB-T (translation-dominant perturbation), GRAB-R (pose-dominant perturbation), and GRAB-B (balanced perturbation). Our model is only trained on the GRAB-B dataset and tested on the five perturbation tasks, as shown in Table~\ref{tab:GRAB}. 

\paragraph{HO3D.}
HO3D \cite{hampali2020honnotate} is a dataset capturing hand-object interaction. It consists of frame-wise annotation for hands and objects with corresponding RGB images. We follow the method in \cite{zhou2022toch} to select a subset of frames containing hand-object interaction. To evaluate our method on real tracking perturbations, we retrained a state-of-the-art hand-object tracking method~\cite{hasson20_handobjectconsist} and applied our method to the predicted results.

\subsection{Evaluation Metrics}
We follow TOCH~\cite{zhou2022toch} to define evaluation metrics as: 
\begin{itemize}
    \setlength\itemsep{0.00em}

    \item \textbf{Mean Per-Joint Position Error (MPJPE)}. MPJPE is the average Euclidean distance between the refined hand joints and the ground truth.
    
    \item \textbf{Mean Per-Vertex Position Error (MPVPE)}. MPVPE is similar to MPJPE which aims to measure the average Euclidean distance between the refined hand vertices and the ground truth vertices.
 
    \item \textbf{Solid Intersection Volume (IV)}. This metric calculates hand-object interpenetration volume by voxelizing the object and hand mesh. Here, we set the voxel size as 2mm. 
    
    \item \textbf{Contact IoU (C-IoU)}. We report the Intersection-over-Union between the refined contact map and the ground truth contact map. To obtain the contact map, we calculate the distance between hand vertices and object mesh and threshold the distance within $\pm$2mm. 
\end{itemize}

\subsection{Implementation Details}
\label{implementation}
Our network consists of three consecutive modules: a hierarchical spatial encoder, a hierarchical temporal encoder, and a reconstruction decoder. The hierarchical spatial encoder consists of four identical blocks with a down-sampling rate of 2. The output dimension of our spatial encoder is $d = 256$. For the Hierarchical Temporal Encoder, we set $d' = d'' = 256$ and set $T = 30$ and $T' = 5$ as the length of input sequence for the long-term temporal transformer and short-term temporal transformer, respectively. Both transformer blocks have four layers and use fixed sine/cosine position encoding. For each layer, the attention head is set as 8, and the feed-forward dimension is set as 1024. 
%----------------------------------------------------Table 3------------------------------------

\begin{table}[!ht]
\small
\centering\setlength{\tabcolsep}{6.4pt}
    \begin{tabular}{l|c|c|c|c}
    \toprule
    Methods & MPJPE &  MPVPE & IV &  C-IoU  \\ \midrule
    Hand Vertices & 7.89 & 8.80 & 1.73 & 24.92 \\
    Object Center & 7.89 & 8.82 & 1.73 & 24.71 \\
    Bounding Box  & 8.02 & 8.92 & 1.75 & 24.91 \\
    Closest Vertex (Ours) & \textbf{7.17} & \textbf{8.09} & \textbf{1.69} & \textbf{26.36} \\
    \bottomrule
    \end{tabular}
    \caption{Quantitative evaluation with different hand-centric representations. For this experiment, we evaluate all the features using the GRAB-B (balanced perturbation) dataset. For a fair comparison, we only replace our representation and keep other settings untouched.} 
    \label{tab:feature}
\end{table}

For the reconstruction decoder, four identical PointNet~\cite{qi2017pointnet} akin blocks are used to map the latent dimension into 16, and a fully connected layer is used to reconstruct the position of each point. 

For all experiments, we set the batch size as 32, and use ADAM as the optimizer with an initial learning rate of $3e^{-4}$, and weight decay of $1e^{-6}$. Training our model takes about 16 hours on 4$\times$NVIDIA Tesla V100S-32G GPUs.

\subsection{Comparison With State-of-the-Arts}
Our goal is to apply the designed method to real hand-object interaction tracking systems or synthesis methods to alleviate the erroneous. However, directly targeting specific tracking or synthesis methods may cause overfitting to specific errors making the performance difficult to quantify. Thus, we evaluate our method and compare it with TOCH \cite{zhou2022toch} on both synthetic tracking dataset and real tracking estimator~\cite{hasson20_handobjectconsist}, where the results are based on their released source codes. 

\paragraph{Quantitative results.}
We show the comparison results in
Table~\ref{tab:GRAB} and Table~\ref{tab:tracking} on GRAB~\cite{taheri2020grab} test set and HO3D~\cite{hampali2020honnotate} test set. For the synthetic GRAB dataset, Our model achieves a significant improvement in MPJPE and MPVPE for all perturbation strategies, which suggests the effectiveness and robustness of our method for refining the incorrect hand pose. For interaction metrics, though there was a marginal diminution in the performance of IV for the GRAB-R dataset, our method achieves notable improvements with C-IoU. For the HO3D dataset, we retrain the previous tracking method~\cite{hasson20_handobjectconsist} as our baseline model and applied both static and dynamic refining methods~\cite{taheri2020grab,grady2021contactopt,zhou2022toch} on the predicted results. As demonstrated in Table~\ref{tab:tracking}, Our method outperforms previous approaches on all three metrics.

\paragraph{Qualitative results.}
The visualization results are presented in Figure~\ref{fig:visualization1} and Figure~\ref{fig:visualization2}. It can be seen that the reconstruction results of TOCH ~\cite{zhou2022toch} suffer from unrealistic pose and incorrect interaction, while our method can generate hand-object interaction motion with better quality.

\subsection{Ablation Study}
For all ablation studies, we report MPJPE, MPVPE, IV, and C-IoU four metrics on the balanced perturbation dataset.

%-------------------------------------------------Table4---------------------------------------
\begin{table}[ht]
\small
\centering\setlength{\tabcolsep}{4.9pt}
    \begin{tabular}{l|c|c|c|c}
    \toprule
    Methods & MPJPE &  MPVPE & IV &  C-IoU  \\ \midrule
    w/o hierarchical-spatial  & 7.35 & 8.22 & 1.94 & 25.25 \\
    short-long-term temporal  & 8.85 & 9.87 & 1.62 & 20.20 \\
    w/o hierarchical-temporal  & 7.47 & 8.39 & 1.72 & 25.70 \\
    w/o post-optimization  & 7.62 & 8.50 & 1.54 & 21.87 \\
    Ours & \textbf{7.17} & \textbf{8.09} & \textbf{1.69} & \textbf{26.36} \\
    \bottomrule
    \end{tabular}
    
    \caption{Quantitative results with various baseline models are presented using the GRAB-B (balanced perturbation) dataset. In this experiment, we evaluate each baseline model by solely replacing the corresponding component while keeping all others unchanged. }
    \label{tab:ablation}
\end{table}

%-----------------------------------------------Table5----------------------------------------

\begin{table}[ht]
\small
\centering\setlength{\tabcolsep}{9.7pt}
    \begin{tabular}{l|c|c|c|c}
    \toprule
    Methods & MPJPE &  MPVPE & IV &  C-IoU \\ \midrule
    Perturbation & 17.82& 18.85 & 2.10 & 12.31 \\
    TOCH & 14.11 & 15.87 & 1.45 & 21.44 \\
    Ours & ~~\textbf{7.80} & ~~\textbf{8.71} & \textbf{1.60} & \textbf{26.14} \\
    \bottomrule
    \end{tabular}
     \caption{Quantitative results on the longer distance setting. In this experiment, we extended the hand-object distance to 30cm, which is twice the distance used in the previous setting. The evaluation was conducted directly using the network parameters trained on the short-distance setting.}
     \label{tab:longer}
\end{table}

\paragraph{Investigating Hand-Centric representation.}
We investigate the effect of the proposed hand-centric representation by comparing it with different hand-object corresponding representations. For the object center representation, we replace the closest vertex with the mass center of the object to compute the interaction feature, while we use eight vertices of the object bounding box for bounding box representations.
The comparative results of different representations are demonstrated in Table~\ref{tab:feature}. It can be seen from the table that our representation achieves the best result on four metrics. The performance of different representations indicates the importance of integrating the relevant object information while effectively preserving the distinct characteristics of each hand vertex.

\paragraph{Investigating each component.} 
We conducted ablation studies to evaluate the impact of each component in our model, and the results are summarized in Table~\ref{tab:ablation}. We investigate the effectiveness of the spatial encoder by replacing our hierarchical spatial encoder with a baseline model using four identical PointNet~\cite{qi2017pointnet} layers. The effectiveness of the hierarchical temporal encoder is examined by swapping long-term and short-term transformers, and by replacing short-term with long-term transformers to validate the hierarchy. Additionally, we also provide a comparison of whether there is post-optimization. Results show the superiority and essentiality of our method.

\paragraph{Experiments on a long hand-object distance.}
We also provide quantitative results for longer-distance hand-object interaction in Table~\ref{tab:longer}. Specifically, we first filter out the motion frames with more than 30cm hand-object distance. Subsequently, we assess the model performance using the models initially trained for short-distance interactions. Our method outperforms TOCH~\cite{zhou2022toch} on all evaluation metrics significantly. Our better performance demonstrates the importance of hand-centric representation. The proposed hand-centric representation can effectively capture hand motions, even when there is a considerable distance between the hand and the object.

%----------------------------------------Conclusion-------------------------------------
\section{Conclusion}
In this paper, we tackle the problem of refining perturbed hand-object interaction sequences. To effectively represent the spatial-temporal interaction sequence, we explore various representations and propose a hand-centric approach that can avoid the ambiguous projection process in the previous object-centric representation. To obtain the dynamic clues of the interaction sequence, we design a hierarchical spatial-temporal architecture. This architecture effectively captures both local and global information. As a result, our method achieves state-of-the-art performance and demonstrates strong generalization capabilities on long-distant interaction sequences.

\section*{Acknowledgements}
This paper is supported by Shandong Excellent Young Scientists Fund Program (Overseas) 2023HWYQ-114.

\bibliography{aaai24}

\end{document}

%% file: math_commands.tex
\usepackage{amsmath,amsfonts,bm}

\def\eqref#1{equation~\ref{#1}}
\def\1{\bm{1}}

\def\vc{{\bm{c}}}

\def\vf{{\bm{f}}}

\def\vh{{\bm{h}}}

\def\vl{{\bm{l}}}

\def\vo{{\bm{o}}}

\def\vr{{\bm{r}}}
\def\vs{{\bm{s}}}

\def\mH{{\bm{H}}}

\def\mL{{\bm{L}}}

\def\mO{{\bm{O}}}

\def\mR{{\bm{R}}}
\def\mS{{\bm{S}}}

\DeclareMathAlphabet{\mathsfit}{\encodingdefault}{\sfdefault}{m}{sl}
\SetMathAlphabet{\mathsfit}{bold}{\encodingdefault}{\sfdefault}{bx}{n}
\newcommand{\tens}[1]{\bm{\mathsfit{#1}}}

\def\tH{{\tens{H}}}

\def\tO{{\tens{O}}}

\def\tR{{\tens{R}}}